\documentclass[sigconf,nonacm]{acmart}

\setcopyright{none}
\acmDOI{}
\acmISBN{}
\usepackage{multirow}
\usepackage{epstopdf}
\usepackage{placeins}
\usepackage{tabularx}
\usepackage{microtype}
\usepackage{subcaption}
\usepackage{array}
\usepackage{makecell}
\usepackage{float}
\newcommand{\cmark}{\ding{51}}
\newcommand{\xmark}{\ding{55}}

\allowdisplaybreaks

\AtBeginDocument{%
  }

\setcopyright{acmlicensed}
\copyrightyear{2027}
\acmYear{2027}
\acmConference[KDD '27]{the 33rd ACM SIGKDD Conference on Knowledge Discovery and Data Mining}{August 2027}{San Jose, CA, USA}
\newcommand{\method}{PGU-Decoder}
\DeclareMathOperator*{\argmax}{arg\,max}

\begin{document}

\title{PGU-Decoder: Prior-Guided Unified Decoding for Industrial Road-Network Map Updating}

\author{Yuguang Ning}
\affiliation{%
  \institution{Amap, Alibaba Group}
  \city{Beijing}
  \country{China}}
\authornote{Corresponding author.}
\email{ningyuguang.nyg@alibaba-inc.com}

\author{Bingbing Kang}
\affiliation{%
  \institution{Amap, Alibaba Group}
  \city{Beijing}
  \country{China}}
\email{kangbingbing.kbb@alibaba-inc.com}

\author{Nan Zhao}
\affiliation{%
  \institution{Amap, Alibaba Group}
  \city{Beijing}
  \country{China}}
\email{zoe.zn@autonavi.com}

\author{Zhixuan Li}
\affiliation{%
  \institution{Amap, Alibaba Group}
  \city{Beijing}
  \country{China}}
\email{lizhixuan.lzx@alibaba-inc.com}

\author{Ze Zhao}
\affiliation{%
  \institution{Amap, Alibaba Group}
  \city{Beijing}
  \country{China}}
\email{zhaoze.zz@alibaba-inc.com}

\author{Dongdong Liu}
\affiliation{%
  \institution{Amap, Alibaba Group}
  \city{Beijing}
  \country{China}}
\email{ldd258873@alibaba-inc.com}

\author{Wenjian Zhu}
\affiliation{%
  \institution{Amap, Alibaba Group}
  \city{Beijing}
  \country{China}}
\email{taisu.zwj@alibaba-inc.com}

\author{Zekai Liu}
\affiliation{%
  \institution{Amap, Alibaba Group}
  \city{Beijing}
  \country{China}}
\email{liuzekai.lzk@alibaba-inc.com}

\author{Junhui Wang}
\affiliation{%
  \institution{Amap, Alibaba Group}
  \city{Beijing}
  \country{China}}
\email{junhui.wang@alibaba-inc.com}


\begin{abstract}
Timely updates of map data, especially road-network map data, are essential to navigation services and autonomous driving, as they directly affect the reliability of route planning, the accuracy of navigation decisions, and the safety of vehicle control. Among the core challenges in road-network updating are intra-intersection road network generation, merging/diverging detection, and drivable-area detection, which correspond to the topology, geometry, and attributes of road-network map data, respectively. Since these tasks are typically handled by separate pipelines, they require duplicated effort in rule maintenance, quality assurance, and manual verification, increasing overall operational costs.
To address these challenges, we propose \method{}, a production-grade unified decoding framework for road-network updating. Within a single decoding framework, \method{} produces semantic predictions for the topology, geometry, and attribute dimensions. Specifically, we design a Prior Query Module (PQM) that constructs point-level prior queries from sampled locations on road-network geometries. We then use fixed reference points throughout decoding, preserving the positional priors and enabling the model to focus on surrounding context. The point-level representations are further aggregated into instance-level representations, while a Relational Graph Attention Network (RGAT)  captures inter-instance dependencies. Finally, task-specific prediction heads generate semantic predictions for the respective update tasks.
We validate \method{} in large-scale automated production of road-network updates spanning more than 360 cities in China. Compared with seven baselines, \method{} consistently outperforms all of them. Following production deployment at AutoNavi Map, \method{} achieves quality scores exceeding 98\% across all three representative tasks, reduces manual verification costs by 95\%, and improves the update frequency from weekly to hourly. These results demonstrate that \method{} is a practical and cost-effective industrial solution for map data updating.
\end{abstract}


% %%
% %% The code below is generated by the tool at http://dl.acm.org/ccs.cfm.
% %% Please copy and paste the code instead of the example below.
% %%




\maketitle

\section{Introduction}
\label{sec:intro}

Digital maps serving one billion users are underpinned by large-scale, structured road-network map data that support route planning, navigation guidance, and driving safety. Our map-update system covers more than 360 cities and 16 million kilometers of roads, handling 360 million user service requests daily. At this scale, even a small error rate can lead to suboptimal route planning, navigation deviations, or incorrect maneuver instructions, and may introduce driving-safety risks. Ensuring map-data correctness and production efficiency under large-scale, rapid updating is therefore a central challenge for industrial-grade map systems.

Industrial road-network map data are organized at two levels of granularity: road-level map data and lane-level map data. Road-level map data represent macroscopic connectivity, supporting feasible route planning and accurate representation of intersection maneuvers. Lane-level map data describe fine-grained driving space and traffic constraints, determining whether vehicles may cross boundaries or enter adjacent areas. In production, lane-level map data are generated through a coarse-to-fine workflow. Crowdsourced data collection first produces coarse lane-level map data, which are subsequently refined through topology construction, geometric correction, and attribute assignment to form high-quality lane-level map data for advanced driver-assistance systems and autonomous driving~\cite{li2023delving}. Both data levels contain topology, geometry, and attributes. Topology characterizes connectivity among road-network elements; geometry describes the spatial configurations of road and lane elements; and attributes encode drivability and other driving constraints. The interplay among data granularity, representation dimensions, and production stages makes scalable road-network updating a multi-level automation challenge.

\begin{figure}[t]
    \centering
    \includegraphics[width=1\linewidth]{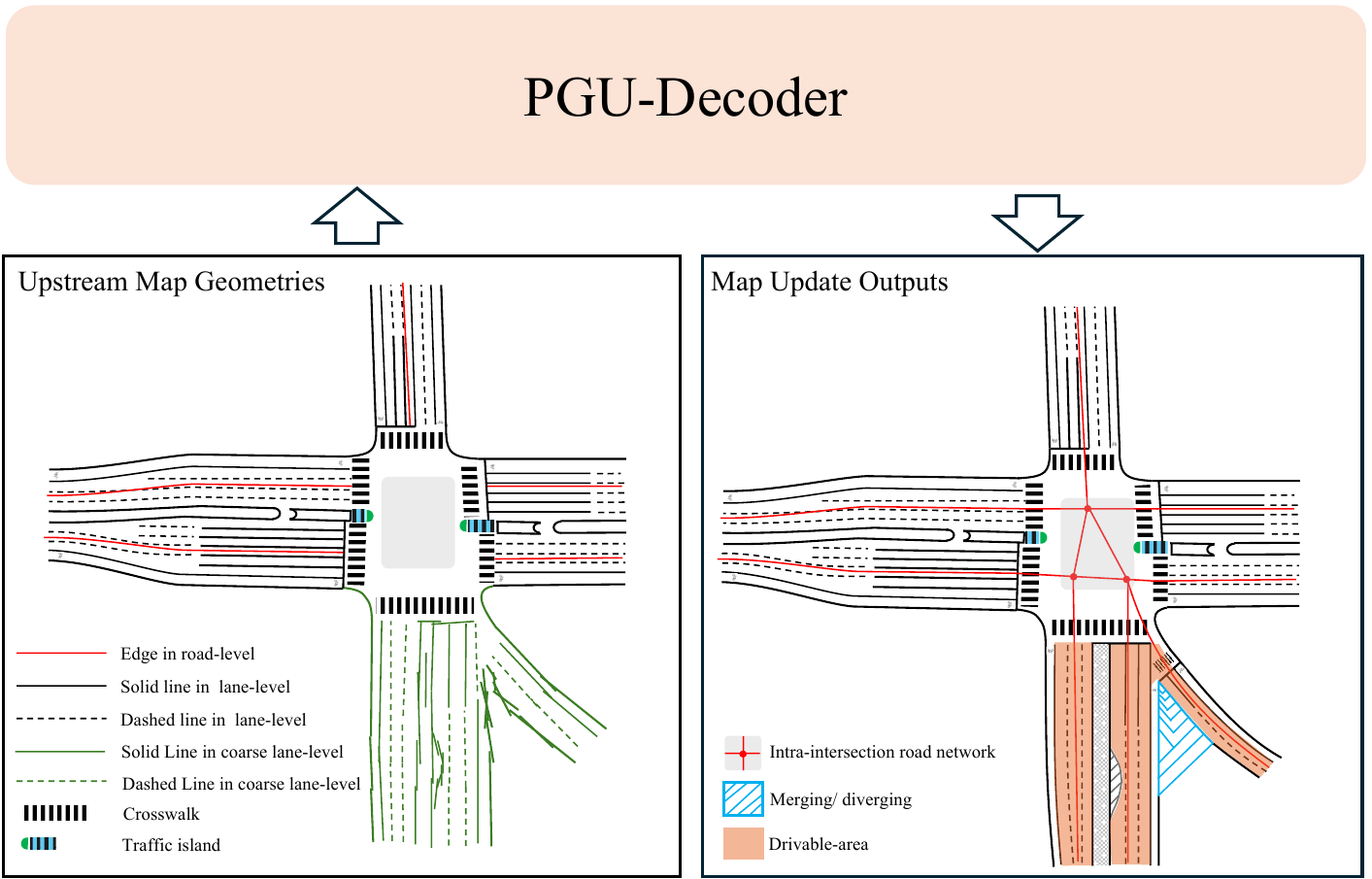}
    \caption{Overview of the three road-network map updating tasks supported by PGU-Decoder. In the output panel, the underlying road-level and high-quality lane-level geometries are supplied by upstream pipelines; colored overlays denote the task-specific outputs.}
    \label{fig:business_dia}
\end{figure}

We study three representative tasks in the road-network updating workflow. Intra-intersection road network generation constructs road-level topology within intersections, supporting route planning and maneuver instructions. Merging/diverging detection determines whether candidate geometries in coarse lane-level map data correspond to actual merging or diverging locations, despite noise such as missing, redundant, or erroneous lane dividers. Drivable-area detection determines the drivability of regions on either side of lane dividers in high-quality lane-level map data. Together, the three tasks address road-level topology, coarse lane-level geometry, and high-quality lane-level attributes, respectively, spanning key stages of road-network data production: establishing reachability, identifying critical geometric structures, and assigning driving constraints. Figure~\ref{fig:business_dia} illustrates the three tasks on a representative intersection, showing the upstream map geometries used as input and the corresponding map-update outputs.

The three tasks present distinct challenges. Intra-intersection road network generation is rule-intensive: road connectivity is jointly constrained by intersection configurations, road classes, turning restrictions, and AutoNavi Map production specifications. More than 400 intersection configurations make manual construction slow and error-prone, with an average manual processing cost of approximately CNY 25 per intersection.
Merging/diverging detection is noise-sensitive, with manual processing costs of approximately CNY 10 per kilometer and consistency that is difficult to ensure.
Drivable-area detection is safety-critical: an incorrect prediction may cause the system to cross a non-drivable boundary or enter an unsafe area. Manual processing costs approximately CNY 12 per kilometer. Together, these tasks represent three core bottlenecks in industrial map updating: rule-intensive topology construction, noise-sensitive geometric recognition, and safety-critical attribute assignment.

Although the three tasks differ in data granularity, production stage, and output semantics, they share a common input: upstream processes provide road-level or lane-level geometries, and the model must make semantic predictions using the surrounding context. Our unification does not rely on joint training or cross-task knowledge transfer; instead, it treats the tasks as structured semantic prediction over upstream geometric priors.
Existing research and production solutions still lack a unified automated framework for semantic predictions. In industrial map-updating operations, tasks are typically developed and operated independently, each with separate workflows, model architectures, and evaluation protocols. This separation duplicates model and system development and increases maintenance costs. Rule-based approaches generalize poorly and are difficult to maintain across cities with diverse road configurations and AutoNavi Map production specifications. Query-based object detection~\cite{carion2020end} and end-to-end mapping methods~\cite{liao2022maptr, li2026graph} typically require queries to learn both element localization and semantic predictions, rather than directly exploiting the geometric priors already available from upstream processes. Vision-language models~\cite{zhang2024vision} primarily target image-text semantic understanding and are not well suited to the accuracy requirements of road-network map updating.

To address these limitations, we propose \method{}, a prior-guided decoding framework for industrial road-network map updating. The Prior Query Module (PQM) encodes sampled point locations from known road-level or lane-level geometries as point-level prior queries. Fixed reference points preserve these positional priors throughout decoding, avoiding layer-wise positional refinement and allowing queries to focus on surrounding context and semantic information rather than element localization. Point-level representations are then aggregated into instance-level representations, and RGAT~\cite{busbridge2019relational} captures inter-instance dependencies. Finally, task-specific output heads generate the respective semantic predictions, allowing all three tasks to be addressed within a single decoding architecture.

We evaluate \method{} offline and deploy it in production across the three tasks and data granularities. Our offline evaluation data are drawn from more than ten representative cities across China, with a controlled proportion of urban and suburban samples to assess generalization. \method{} consistently outperforms the seven baseline methods. Following deployment in the AutoNavi Map production system, \method{} achieves quality scores exceeding 98\% on all three tasks, reduces manual verification costs by 95\%, and raises the update frequency from weekly to hourly. Our main contributions are as follows:
\begin{itemize}
    \item \textbf{Unified problem formulation.} We formulate the three tasks as structured semantic prediction problems over upstream geometric priors, spanning the topology, geometry, and attributes of road-network map data.

    \item \textbf{Unified decoding framework.} We propose \method{}, which combines PQM, fixed reference points, and RGAT to support intra-intersection road network generation, merging/diverging detection, and drivable-area detection within a single decoding architecture.

    \item \textbf{Production deployment validation.} Large-scale offline evaluations and production deployment demonstrate the effectiveness of \method{} for industrial road-network map updating.
\end{itemize}

\section{Related Work}
Existing map-extraction research spans remote-sensing road mapping and onboard lane perception. Remote-sensing methods recover road geometry and connectivity through segmentation post-processing, graph search, or sequential generation~\cite{mattyus2017deeproadmapper,chu2019neural,liu2018roadnet,bastani2018roadtracer,he2020sat2graph,xu2022rngdet}; onboard approaches have progressed from pixel-level segmentation to parameterized and instance-level modeling in 3D or bird's-eye-view (BEV) spaces~\cite{neven2018towards,liu2021condlanenet,garnett20193d,guo2020gen,liu2021end,chen2022persformer}. Recent end-to-end methods generate vectorized map geometry~\cite{li2022hdmapnet,liu2023vectormapnet,liao2022maptr,liao2025maptrv2,qiao2023end,ding2023pivotnet,yuan2024streammapnet}, while structured methods model topological relations among map elements~\cite{can2021structured,xu2022centerlinedet,li2026graph,wu2024topomlp,li2024lanesegnet,fu2024topologic}. OpenLane-V2~\cite{wang2023openlane} provides a standardized benchmark, and LaneGAP~\cite{liao2024lane} and SMERF~\cite{luo2024augmenting} study lane-graph generation and map-prior utilization. At industrial scale, DuMapNet~\cite{xia2024dumapnet} generates city-scale lane-level maps from BEV, and LDMapNet-U~\cite{xia2025ldmapnet} extends it to map updating. Regardless of whether historical maps are additionally used as auxiliary references, these methods still infer map geometry directly from raw sensor inputs, making downstream reasoning susceptible to localization errors, missed detections, and redundant predictions. In contrast, our setting starts from road-level or lane-level geometries provided upstream and formulates the three update tasks as semantic predictions over existing geometries.  A prior-query formulation was previously introduced for merging/diverging detection~\cite{ning2025prior}; PGU-Decoder generalizes it into a unified Prior Query Module that additionally covers intra-intersection road-network generation and drivable-area detection.

DETR~\cite{carion2020end} introduces learnable object queries for end-to-end detection. Deformable DETR~\cite{zhu2020deformable} uses reference points to guide attention sampling and iteratively refines locations across decoder layers; DAB-DETR~\cite{liu2022dab} represents queries as anchor boxes, and MapTR~\cite{liao2022maptr} uses instance-level and point-level queries for vectorized map elements. These designs localize previously unknown objects and therefore learn or refine their locations during decoding. In contrast, we encode sampled locations on known road-level or lane-level geometries as point-level prior queries and retain their reference points throughout decoding, allowing queries to focus on surrounding context and inter-instance relations.

\section{Method}

\begin{figure*}[t]
    \centering
    \includegraphics[width=0.9\linewidth]{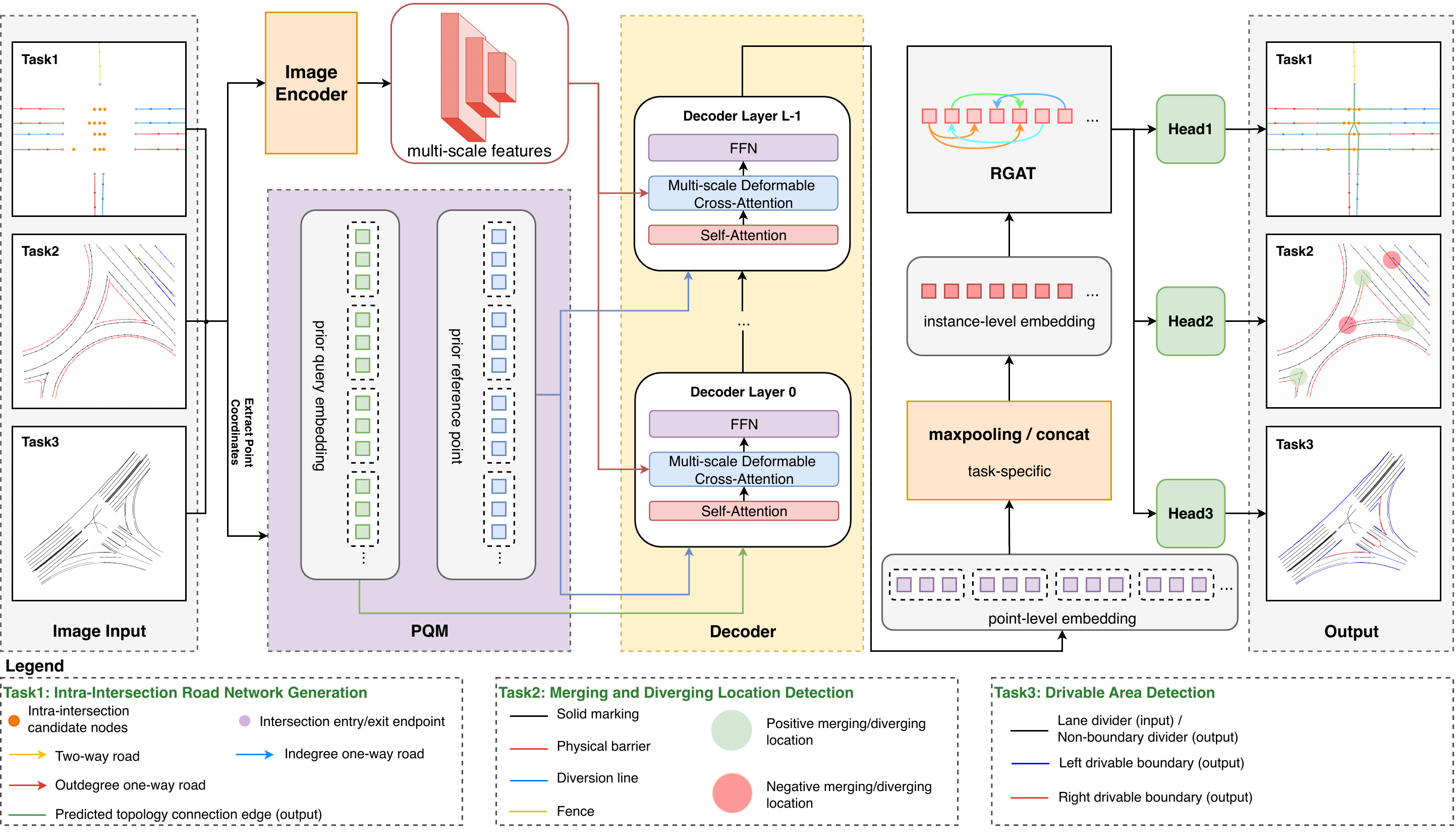}
    \caption{Framework of the \method{}.}
    \label{fig:illustration}
\end{figure*}

\begin{figure}[htbp]
  \centering
  \begin{subfigure}[t]{0.49\linewidth}
    \centering
    \includegraphics[width=\linewidth]{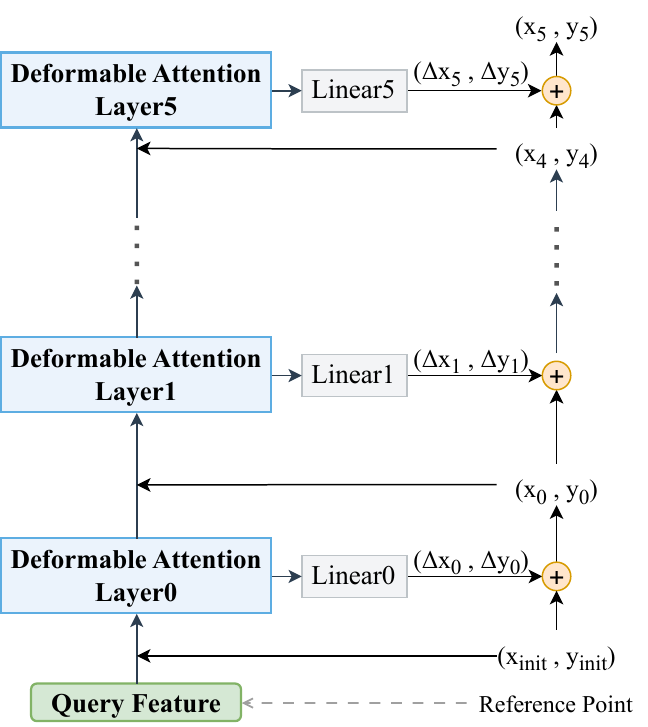}
    \caption{Learnable reference points}
    \label{fig:ref_point_a}
  \end{subfigure}
  \hfill
  \begin{subfigure}[t]{0.49\linewidth}
    \centering
    \includegraphics[width=\linewidth]{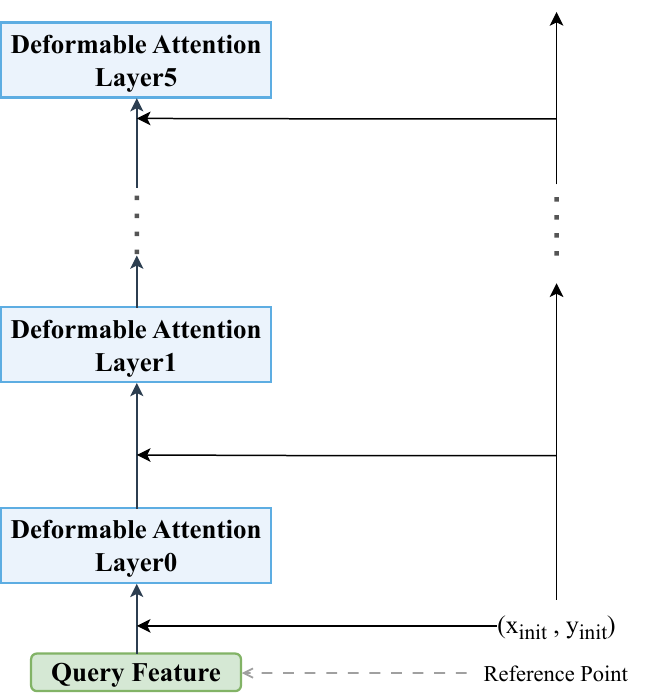}
    \caption{Fixed reference points}
    \label{fig:ref_point_b}
  \end{subfigure}
  \caption{Comparison of reference-point mechanisms. }
  \label{fig:ref_point_compare}
\end{figure}

\begin{figure*}[t]
  \centering
      \begin{subfigure}[t]{0.33\textwidth}
      \centering
      \includegraphics[height=4.3cm]{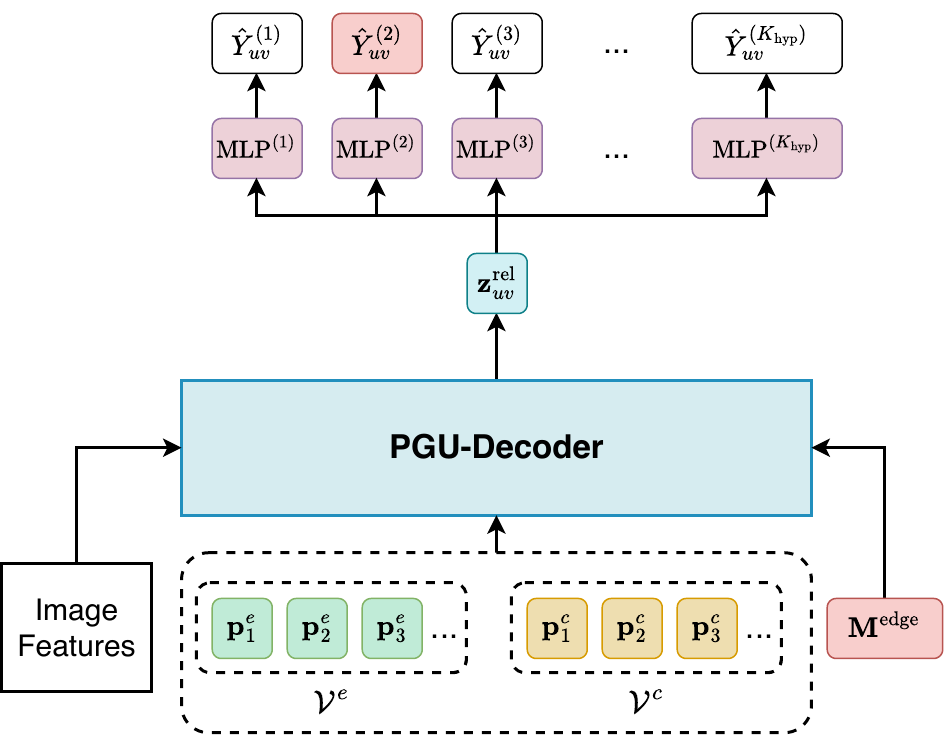}
      \caption{Intra-intersection road network generation}
      \label{fig:arch_comparison_a}
    \end{subfigure}
    \hfill
    \begin{subfigure}[t]{0.33\textwidth}
      \centering
      \includegraphics[height=4.3cm]{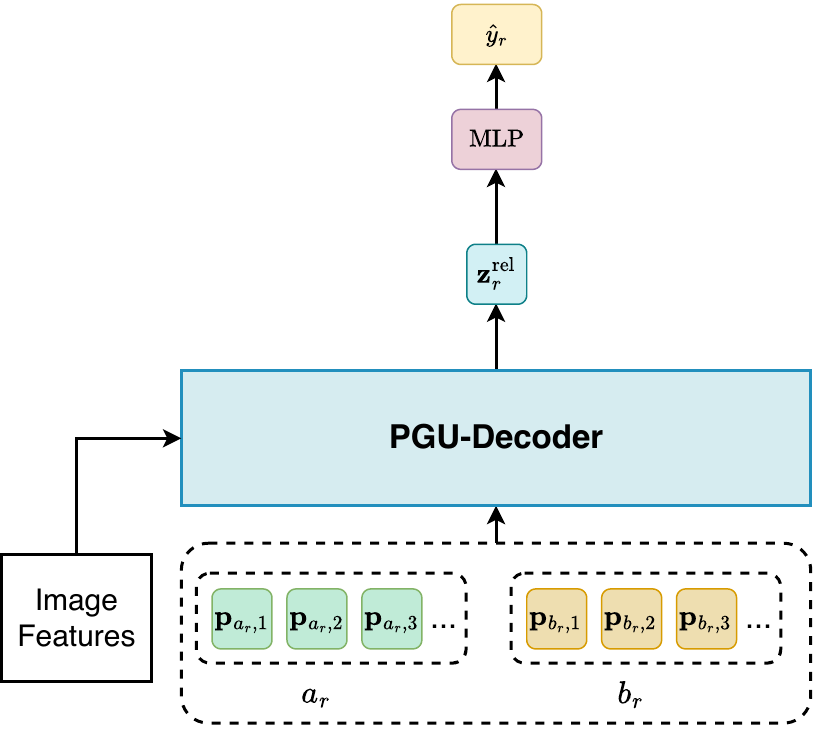}
      \caption{Merging/diverging detection}
      \label{fig:arch_comparison_b}
    \end{subfigure}
    \hfill
    \begin{subfigure}[t]{0.33\textwidth}
      \centering
      \includegraphics[height=4.3cm]{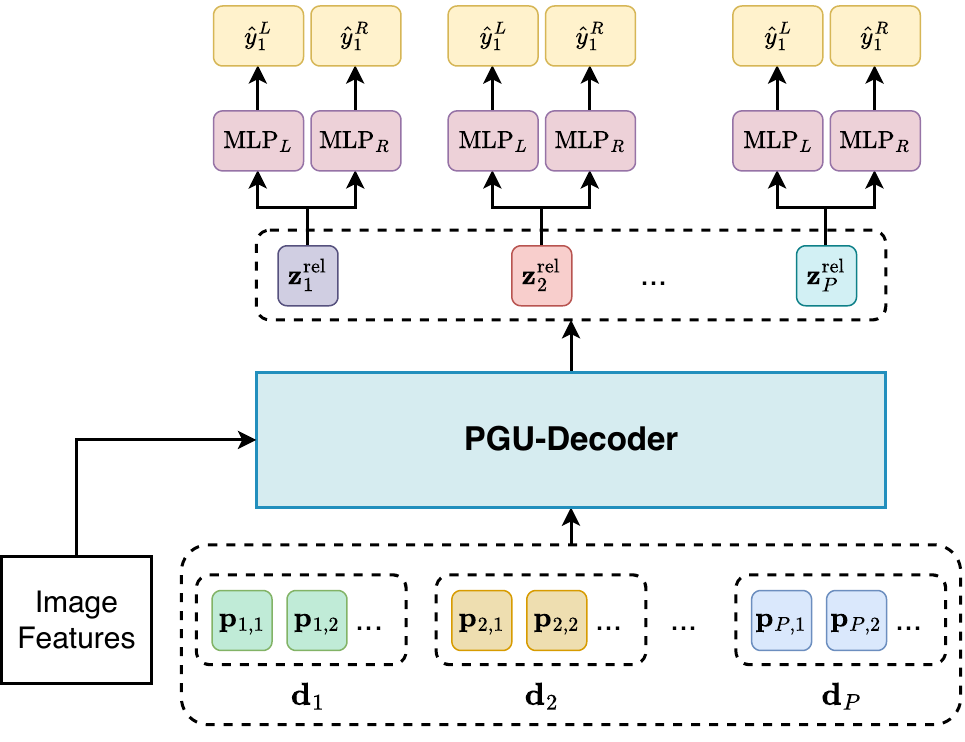}
      \caption{Drivable-area detection}
      \label{fig:arch_comparison_c}
    \end{subfigure}
  \caption{Task-specific formulations of \method{} for the three update tasks.}
  \label{fig:arch_comparison}
\end{figure*}

\subsection{Framework Overview}
As illustrated in Figure~\ref{fig:illustration}, \method{} is a prior-guided unified decoding framework built upon Deformable DETR. It comprises a Prior Query Module (PQM), a fixed reference point mechanism, and a Relational Graph Attention Network (RGAT). For each task, the framework takes road-level or lane-level geometries and their orientations as input, rasterizes them into a bird's-eye-view (BEV) representation, and extracts multi-scale features using a backbone and encoder. The PQM constructs point-level prior queries and their corresponding reference points from sampled geometry locations. Unlike Deformable DETR, which iteratively updates reference points across decoder layers, \method{} keeps the geometry-initialized reference points fixed throughout decoding. This eliminates layer-wise positional refinement and allows the queries to focus on modeling semantics in the surrounding context. The decoder produces point-level embeddings, which are aggregated into instance-level embeddings through task-specific max pooling or concatenation. RGAT then models inter-instance dependencies among neighboring instances. Finally, task-specific prediction heads generate the corresponding semantic predictions. The three tasks instantiate the same prior-guided decoding architecture but are trained independently. The geometric objects anchored by the queries, target semantics, point-level aggregation strategies, and prediction heads are task-specific. The following sections detail the prior-guided decoding modules and the task-specific formulations.
% The three tasks use the same prior-guided decoding framework; 

\subsection{Prior-Guided Decoding}

\subsubsection{Prior Query Module}

The Prior Query Module (PQM) constructs point-level prior queries and corresponding reference points from sampled road-level or lane-level geometries. Let \(\mathbf{p}_{i,j}=(x_{i,j},y_{i,j})\) denote the \(j\)-th sampled point of the \(i\)-th geometric instance.

Given the BEV rendering range \([x_{\min},x_{\max}]\times[y_{\min},y_{\max}]\), the PQM maps each sampled point to normalized BEV coordinates \((\tilde{x}_{i,j},\tilde{y}_{i,j})\in[0,1]^2\). Because the vector and image coordinate systems use opposite \(y\)-axis directions, the \(y\)-coordinate is inverted:
\begin{equation}
(\tilde{x}_{i,j},\tilde{y}_{i,j})
=
\left(
\frac{x_{i,j}-x_{\min}}{x_{\max}-x_{\min}},
\frac{y_{\max}-y_{i,j}}{y_{\max}-y_{\min}}
\right).
\end{equation}
We then apply sinusoidal position encoding (SinPE) to the normalized coordinates to obtain the position embedding:
\begin{align}
\mathbf{e}^{\mathrm{pos}}_{i,j}
&=
\mathrm{SinPE}(\tilde{x}_{i,j},\tilde{y}_{i,j}).
\end{align}
We then define the point-level prior query as the sum of this position embedding and a zero content embedding:
\begin{align}
\mathbf{q}^{\mathrm{pt}}_{i,j}
&=
\mathbf{e}^{\mathrm{pos}}_{i,j}
+
\mathbf{e}^{c},
\qquad
\mathbf{e}^{c}=\mathbf{0},
\end{align}
where \(\mathbf{e}^{c}\) is a zero content embedding. Thus, the initial prior of each point-level query is provided entirely by the locations of the upstream geometries.

The same normalized coordinates are directly used as the corresponding reference point:
\begin{align}
\mathrm{ref}_{i,j}^{(0)}
=
(\tilde{x}_{i,j},\tilde{y}_{i,j}).
\end{align}

Crucially, we treat the BEV image as a feature map at the input resolution. The normalized coordinates both locate the geometry in the BEV image and align with the reference-point coordinates of feature maps at different scales, thereby serving as prior reference points. The PQM therefore converts existing geometries into explicit positional priors for the decoder.

\subsubsection{Fixed Reference Points}

As illustrated in Figure~\ref{fig:ref_point_a}, Deformable DETR iteratively refines reference points across decoder layers. Starting from initial reference points, each layer predicts coordinate offsets and passes the updated reference points to the next layer. This design is appropriate when object locations must be inferred from the input.

In our setting, the PQM initializes each query with a reference point derived from upstream road-level or lane-level geometries. As shown in Figure~\ref{fig:ref_point_b}, \method{} reuses this initial reference point throughout the decoder:
\begin{equation}
\mathrm{ref}_{i,j}^{(\ell)}
=
\mathrm{ref}_{i,j}^{(0)},
\qquad
\ell=0,\ldots,L-1,
\end{equation}
where \(L\) is the number of decoder layers and \(\mathrm{ref}_{i,j}^{(0)}\) is the prior reference point constructed by PQM. The decoder can therefore focus on semantic information in the surrounding context instead of repeatedly refining the reference points. Fixing reference points only fixes the sampling centers; deformable attention still learns local sampling offsets around them.

\subsubsection{Relational Graph Attention Network}

RGAT takes task-specific instance-level embeddings \(\mathbf{z}_{i}\) as input. Each \(\mathbf{z}_{i}\) is obtained by aggregating the point-level embeddings produced by the decoder. 
For each task, a task-specific buffer constructs a relation matrix from upstream map information available before prediction. For instances \(i\) and \(k\), \(M_{ik}^{\mathrm{rel}} \in \{0,1,\ldots,R\}\) encodes the discrete type of their relation derived from upstream map information. The nonzero values may represent additional task-specific relation types. If no valid relation exists, \(M_{ik}^{\mathrm{rel}}=0\), and the pair is masked. The relation matrix never uses ground-truth task labels or target structures. In particular, for intra-intersection road network generation, it contains no connectivity within the intersection, which is the target of prediction.
For a target instance \(i\), let
\[
\mathcal{N}(i)
=
\left\{k \mid M_{ik}^{\mathrm{rel}}\neq 0\right\}.
\]
denote the related instances specified by the relation matrix. By construction, \(\mathcal{N}(i)\neq\varnothing\) for every instance processed by RGAT. For each \(k\in\mathcal{N}(i)\), the relation entry \(M_{ik}^{\mathrm{rel}}\) is mapped to a learned relation embedding \(\mathbf{e}_{ik}^{\mathrm{rel}}\).

% For each task, a task-specific buffer constructs a relation matrix from information available in the upstream map before prediction. For instances \(i\) and \(k\), a nonzero entry indicates a predefined relation derived from known map geometry and geometric proximity. otherwise, the pair is masked. The relation matrix never uses task labels or target structures. For a target instance \(i\), let \(\mathcal{N}(i) \neq \emptyset\) denote its related instances specified by the relation matrix. The relation between \(i\) and \(k\in\mathcal{N}(i)\) is mapped to a learned relation embedding \(\mathbf{e}_{ik}^{\mathrm{rel}}\).

% For instances \(i\) and \(k\), \(M_{ik}^{\mathrm{rel}}=1\) if they are connected by known upstream map relations or fall within the task-specific spatial buffer; otherwise, \(M_{ik}^{\mathrm{rel}}=0\), and the pair is masked.

% For instances \(i\) and \(k\), a nonzero entry indicates a predefined relation derived from known map geometry and geometric proximity; otherwise, the pair is masked. 

% For instances \(i\) and \(k\), \(M_{ik}^{\mathrm{rel}} \in \{0,1,\ldots,R\}\) encodes the discrete type of their relation derived from upstream map information. Other nonzero values may represent additional task-specific relation types. If no valid relation exists, \(M_{ik}^{\mathrm{rel}}=0\), and the pair is masked. Each nonzero relation type is mapped to a distinct learnable relation embedding \(\mathbf e_{ik}^{\mathrm{rel}}=\operatorname{Emb}(M_{ik}^{\mathrm{rel}})\).

For a single attention head, the relation embedding contributes to both attention scores and value aggregation:
\begin{align}
s_{ik}
&=
\frac{
(\mathbf{W}_{Q}\mathbf{z}_{i})^{\top}
\left(
\mathbf{W}_{K}\mathbf{z}_{k}
+
\alpha\mathbf{W}_{K}^{\mathrm{rel}}
\mathbf{e}_{ik}^{\mathrm{rel}}
\right)
}
{\sqrt{d_h}}, \\
a_{ik}
&=
\frac{\exp(s_{ik})}
{\sum_{k'\in\mathcal{N}(i)}\exp(s_{ik'})}, \\
\mathbf{c}_{i}
&=
\sum_{k\in\mathcal{N}(i)}
a_{ik}
\left(
\mathbf{W}_{V}\mathbf{z}_{k}
+
\beta\mathbf{W}_{V}^{\mathrm{rel}}
\mathbf{e}_{ik}^{\mathrm{rel}}
\right).
\end{align}
where \(\mathbf{W}_{Q}\), \(\mathbf{W}_{K}\), and \(\mathbf{W}_{V}\) are the query, key, and value projection matrices, respectively; \(\mathbf{W}_{K}^{\mathrm{rel}}\) and \(\mathbf{W}_{V}^{\mathrm{rel}}\) project relation embeddings for attention scoring and value aggregation; \(d_h\) is the dimension of one attention head; and \(\alpha=1\) and \(\beta=1\) control the contributions of relation embeddings. Thus, \(\mathbf{c}_{i}\) aggregates their instance and relation information.

The outputs from all heads are concatenated and linearly projected. Standard residual and feed-forward components further refine the resulting embeddings. Stacking multiple RGAT layers produces relation-aware instance embeddings for the task-specific prediction heads.

\subsection{Task-Specific Formulations and Prediction Heads}

\subsubsection{Intra-Intersection Road Network Generation}

Figure~\ref{fig:arch_comparison_a} illustrates the overall pipeline of this task. Let \(\mathcal{S}\) denote the directed road segments incident to an intersection, where each segment is represented as a polyline whose vertex order specifies the driving direction. For each segment, we select the endpoint nearest to the intersection boundary, yielding the set of entry/exit endpoints \(\mathcal{V}^{e}\). As the input map lacks intra-intersection topology, we introduce internal transit-node candidates \(\mathcal{V}^{c}\) to enable recovering the missing road connections.

\begin{figure}[htbp]
  \centering
  \includegraphics[width=0.4\linewidth]{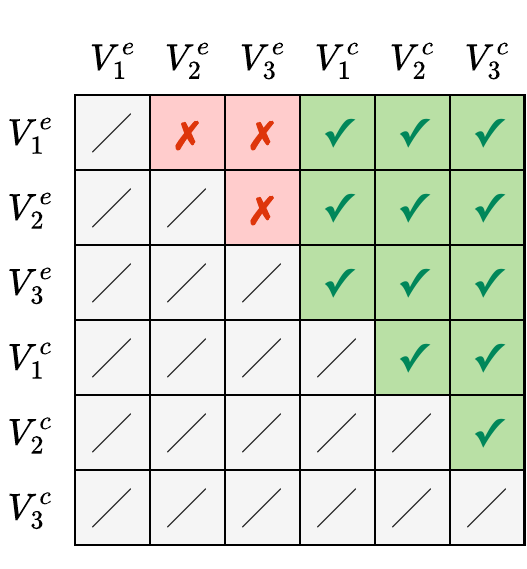}
  \caption{Edge-prediction mask for intra-intersection road network generation. Green cells denote evaluated node pairs, and red cells denote excluded direct entry/exit pairs.}
  \label{fig:edge_mask}
\end{figure}

We generate \(\mathcal{V}^{c}\) by extending the incident road segments into the intersection and intersecting the extensions with lines between entry/exit endpoints. After clustering and merging the resulting intersection points, we retain at most \(K_c=20\) transit-node candidates:
\begin{align}
|\mathcal{V}^{c}|
\leq
K_c,
\qquad
K_c=20.
\end{align}
The complete node set is
\begin{align}
\mathcal{V}
=
\mathcal{V}^{e}
\cup
\mathcal{V}^{c}.
\end{align}
We construct an edge-prediction mask \(\mathbf{M}^{\mathrm{edge}}\) (Figure~\ref{fig:edge_mask}), where \(M_{uv}^{\mathrm{edge}}=1\) marks a node pair evaluated by the prediction head. Because the direction of a recovered road connection is determined by its entry and exit endpoints, we predict undirected connectivity and evaluate each unordered pair once, avoiding redundant directional edge decisions. The mask excludes impermissible pairs, including direct entry/exit endpoint pairs.

For each node \(u\in\mathcal{V}\), its coordinate is used as a single sampled point by PQM to construct a point-level prior query. The decoder outputs the corresponding point-level embedding \(\mathbf{h}_{u}^{\mathrm{pt}}\). For each node pair selected by the edge-prediction mask, we construct a pair embedding:
\begin{equation}
\mathbf{z}_{uv}
=
\operatorname{concat}
\left(
\mathbf{h}_{u}^{\mathrm{pt}},
\mathbf{h}_{v}^{\mathrm{pt}}
\right),
\qquad
M_{uv}^{\mathrm{edge}}=1.
\end{equation}
Each \(\mathbf{z}_{uv}\) serves as a task-specific instance-level embedding and is processed by RGAT. The resulting relation-aware pair embedding is denoted by \(\mathbf{z}_{uv}^{\mathrm{rel}}\).

To account for multiple feasible connectivity patterns, we employ a multi-hypothesis prediction head comprising \(K_{\mathrm{hyp}}=4\) independent MLPs. The \(m\)-th head predicts the probability that node pair \((u,v)\) corresponds to a road connection:
\begin{align}
\hat{Y}_{uv}^{(m)}
=
\sigma
\left(
\operatorname{MLP}^{(m)}
\left(
\mathbf{z}_{uv}^{\mathrm{rel}}
\right)
\right),
\qquad
m=1,\ldots,K_{\mathrm{hyp}},
\end{align}
where \(\sigma(\cdot)\) denotes the sigmoid function. Given the ground-truth adjacency label \(Y_{uv}^{\ast}\), we optimize the prediction head using a winner-take-all objective:
\begin{align}
\mathcal{L}_{\mathrm{road}}
=
\min_{m\in\{1,\ldots,K_{\mathrm{hyp}}\}}
\frac{
\sum_{u,v}
M_{uv}^{\mathrm{edge}}
\mathcal{L}_{\mathrm{BCE}}
\left(
\hat{Y}_{uv}^{(m)},
Y_{uv}^{\ast}
\right)
}{
\sum_{u,v}M_{uv}^{\mathrm{edge}}
}.
\end{align}
At inference, we select the hypothesis with the highest mean per-edge decision confidence over valid node pairs:
\begin{align}
m^{\star}
=
\argmax_{m\in\{1,\ldots,K_{\mathrm{hyp}}\}}
\frac{
\sum_{u,v} M_{uv}^{\mathrm{edge}}
\max\!\left(
\hat{Y}_{uv}^{(m)},
1-\hat{Y}_{uv}^{(m)}
\right)
}{
\sum_{u,v}M_{uv}^{\mathrm{edge}}
}.
\end{align}
The final connectivity prediction is
\(\hat{Y}_{uv}=\hat{Y}_{uv}^{(m^{\star})}\). This label-free selection rule prefers the hypothesis with the most confident binary decisions over valid candidate edges.

\subsubsection{Merging/Diverging Detection}

As illustrated in Figure~\ref{fig:arch_comparison_b}, merging/diverging detection operates on coarse lane-level map data, in which lane dividers may be missing, redundant, or geometrically inaccurate. Building on our prior work~\cite{ning2025prior}, we first concatenate head-to-tail connected lane dividers into maximal continuous chains. For each chain endpoint, we retrieve nearby chains that either intersect the current chain or are geometrically likely to intersect it. We crop a \(10\,\mathrm{m}\) segment from each involved chain endpoint, and use the resulting segment pair as a recalled candidate for merging/diverging detection.

Let \((a_r,b_r)\) denote the \(r\)-th recalled pair of cropped lane divider segments. We uniformly sample \(P\) points from each segment and use them to instantiate point-level prior queries through PQM. Let \(\mathbf{h}_{a_r,j}^{\mathrm{pt}}\) and \(\mathbf{h}_{b_r,j}^{\mathrm{pt}}\) denote the decoder outputs for the \(j\)-th sampled point on the two segments. We aggregate point-level embeddings within each segment by max pooling, and concatenate the resulting segment embeddings to form the task-specific instance-level embedding:
\begin{align}
\mathbf{z}_{r}
=
\operatorname{concat}
\left(
\operatorname{MaxPool}_{j=1}^{P}
\left(
\mathbf{h}_{a_r,j}^{\mathrm{pt}}
\right),
\operatorname{MaxPool}_{j=1}^{P}
\left(
\mathbf{h}_{b_r,j}^{\mathrm{pt}}
\right)
\right).
\end{align}
The pair embeddings are processed by RGAT, yielding relation-aware embeddings \(\mathbf{z}_{r}^{\mathrm{rel}}\). A binary prediction head determines whether the recalled pair corresponds to a merging or diverging location:
\begin{align}
\hat{y}_{r}
=
\sigma
\left(
\operatorname{MLP}_{\mathrm{md}}
\left(
\mathbf{z}_{r}^{\mathrm{rel}}
\right)
\right).
\end{align}

\subsubsection{Drivable-Area Detection}

As illustrated in Figure~\ref{fig:arch_comparison_c}, drivable-area detection operates on high-quality lane-level map data. Each oriented lane divider defines a left side and a right side according to its vertex order, and the task predicts whether the region on each side is drivable.

For each lane divider \(\mathbf{d}_i\), we uniformly sample \(P\) points and use PQM to construct a point-level prior query for each sampled point. The prior-guided decoder produces the corresponding point-level embeddings \(\{\mathbf{h}_{i,j}^{\mathrm{pt}}\}_{j=1}^{P}\), which are max-pooled to form the instance-level embedding:
\begin{align}
\mathbf{z}_{i}
=
\operatorname{MaxPool}_{j=1}^{P}
\left(
\mathbf{h}_{i,j}^{\mathrm{pt}}
\right).
\end{align}
RGAT then produces the relation-aware embedding \(\mathbf{z}_{i}^{\mathrm{rel}}\). Two independent binary classifiers predict the drivability of the left and right sides:
\begin{equation}
\hat{y}_{i}^{L}
=
\sigma\!\left(\operatorname{MLP}_{L}\!\left(\mathbf{z}_{i}^{\mathrm{rel}}\right)\right),
\quad
\hat{y}_{i}^{R}
=
\sigma\!\left(\operatorname{MLP}_{R}\!\left(\mathbf{z}_{i}^{\mathrm{rel}}\right)\right).
\end{equation}

\section{Experiment}
In this section, we evaluate the performance of the proposed \method{} extensively.

\subsection{Experimental Setup}

\subsubsection{Datasets}

% No public benchmark covers the road-level, coarse lane-level, and high-quality lane-level map data required by our three tasks. We therefore construct three task-specific datasets using AutoNavi Map data from 12 cities: Beijing, Shanghai, Guangzhou, Shenzhen, Hangzhou, Nanjing, Chongqing, Dongguan, Huizhou, Suzhou, Langfang, and Wuhan. For each task, we use controlled urban-to-suburban sampling ratios and split spatially disjoint map tiles into training, validation, and test sets. Further details are provided in Appendix~\ref{sec:appendix-datasets} (Table~\ref{tab:dataset_stats}).

No public benchmark covers the road-level, coarse lane-level, and high-quality lane-level map data required by our three tasks. We therefore construct three task-specific datasets using AutoNavi Map data from 12 cities. Eight cities are used for training and validation, while the remaining four are held out exclusively for testing. This city-disjoint split evaluates generalization to unseen cities. Further dataset statistics and the city split are provided in Appendix~\ref{sec:appendix-datasets} (Table~\ref{tab:dataset_stats}).

\subsubsection{Metrics}

The three tasks have different output structures; we therefore use task-specific metrics that reflect their production-quality requirements.

\noindent\textbf{Intra-Intersection Road Network Generation.}
We use Intersection Accuracy (I-Acc), which requires the predicted road connections of an intersection to exactly match its ground-truth topology:
\begin{align}
\mathrm{I\text{-}Acc}
=
\frac{1}{|\mathcal{J}|}
\sum_{\mathcal{I}\in\mathcal{J}}
\mathbb{I}
\left[
\hat{\mathcal{E}}_{\mathcal{I}}
=
\mathcal{E}_{\mathcal{I}}
\right],
\end{align}
where \(\mathcal{J}\) is the test set of intersections, \(\mathcal{E}_{\mathcal{I}}\) and \(\hat{\mathcal{E}}_{\mathcal{I}}\) denote the ground-truth and predicted road connections of intersection \(\mathcal{I}\), respectively, and \(\mathbb{I}[\cdot]\) is the indicator function.

\noindent\textbf{Merging/Diverging Detection.}
We evaluate the complete pipeline, including candidate-pair retrieval and semantic classification, using end-to-end Precision (P) and Recall (R) on the final predicted merging/diverging locations. Precision measures the reliability of predicted locations, whereas Recall measures the fraction of ground-truth locations recovered by the complete pipeline. We additionally report candidate coverage, defined as the proportion of ground-truth locations represented by at least one retrieved candidate pair.

\noindent\textbf{Drivable-Area Detection.}
We use Pair Accuracy (P-Acc), which requires the drivability attributes on both sides of a lane divider to be correct:
\begin{align}
\mathrm{P\text{-}Acc}
=
\frac{1}{|\mathcal{D}|}
\sum_{d\in\mathcal{D}}
\mathbb{I}
\left[
(\hat{y}_{d}^{L},\hat{y}_{d}^{R})
=
(y_{d}^{L},y_{d}^{R})
\right],
\end{align}
where \(\mathcal{D}\) is the test set of lane dividers, and \(y_{d}^{L}\) and \(y_{d}^{R}\) denote the ground-truth drivability attributes on the left and right sides of divider \(d\), respectively.

\subsubsection{Baselines}

For the three tasks, we compare \method{} with seven representative baselines. End-to-end map generation methods (e.g., MapTR, DuMapNet~\cite{xia2024dumapnet}, LDMapNet-U~\cite{xia2025ldmapnet}) rely on raw sensor inputs, whereas \method{} takes only upstream geometries as input, making them incompatible with our task setting.

\begin{itemize}
    \item \textbf{Rule-based.}
    We implement conventional geometric rules based on lane-divider geometry for merging/diverging detection.

    \item \textbf{GAT}~\cite{velivckovic2017graph}.
    We apply a Graph Attention Network (GAT) to the constructed intersection node set to predict road connections without BEV features.

    \item \textbf{YOLOv11}~\cite{khanam2024yolov11}.
    We adapt YOLOv11, a CNN-based vision-only detector, for merging/diverging detection and drivable-area detection using BEV raster inputs.
    
    \item \textbf{DINO}~\cite{zhang2022dino}.
    We adapt DINO, a transformer-based detector with learnable queries, for the same two tasks. It infers targets from BEV raster inputs without explicitly encoding upstream vector geometries.

    \item \textbf{Qwen3-VL}~\cite{bai2025qwen3}.
    We fine-tune Qwen3-VL with image and text inputs for the applicable tasks. This baseline evaluates whether multimodal semantic reasoning can support structured road-network map updating.

   \item \textbf{Zero-shot}~\cite{radford2019language}.
    We apply Gemini-3.1-Pro directly to each task without task-specific training or in-context examples.
    
    \item \textbf{Few-shot}~\cite{brown2020language}.
    We provide annotated in-context examples to Gemini-3.1-Pro without updating its parameters.
\end{itemize}

\subsubsection{Implementation Details}
Our method is built on Deformable DETR and trained on four V100 GPUs for 25 epochs. We use the Adam optimizer with an initial learning rate of \(1\times10^{-4}\), reduced by a factor of 10 at epoch 20.

\subsection{Main Results}
Table~\ref{tab:method_comparison} compares \method{} with all applicable baselines on the three update tasks. \method{} achieves the best results on every reported metric, covering road-level topology, coarse lane-level geometry, and high-quality lane-level attributes. In particular, it surpasses GAT by \(+5.52\%\) in I-Acc. The constructed candidate nodes and edge mask can represent the ground-truth topology for approximately \(99\%\) of test intersections. Therefore, candidate coverage is not the primary limitation of the reported I-Acc. For merging/diverging detection, the candidate-retrieval stage covers approximately \(99.5\%\) of ground-truth locations, and the reported P and R are computed end-to-end on the final pipeline outputs.

\begin{table}[htbp]
\centering
\caption{Performance comparison across three road-network map updating tasks.}
\label{tab:method_comparison}
\small
\renewcommand{\arraystretch}{1}
\begin{tabular}{
  l
  @{\hspace{8pt}} c
  @{\hspace{12pt}} c @{\hspace{10pt}} c
  @{\hspace{12pt}} c
}
\toprule
\multirow{2}{*}{Method}
  & \makecell[c]{Intra-\\Intersection}
  & \multicolumn{2}{c}{\makecell[c]{Merging/ \\Diverging}}
  & \makecell[c]{Drivable-\\Area} \\
\cmidrule(lr){2-2} \cmidrule(lr){3-4} \cmidrule(lr){5-5}
  & I-Acc
  & P & R
  & P-Acc \\
\midrule
Rule-based & ---   & 77.88 & 86.67 & ---   \\
GAT        & 92.60 & ---   & ---   & ---   \\
YOLOv11    & ---   & 95.29 & 91.69 & 91.44 \\
DINO       & ---   & 94.03 & 88.97 & 74.74 \\
Qwen3-VL   & 68.00 & 92.68 & 94.74 & 56.37 \\
Zero-shot  & 43.43 & 69.23 & 83.64 & 56.11 \\
Few-shot   & 51.55 & 72.73 & 80.00 & 52.42 \\
\midrule
\textbf{Ours} & \textbf{98.12} & \textbf{98.94} & \textbf{95.64} & \textbf{99.68} \\
\bottomrule
\end{tabular}
\par\smallskip
\begin{minipage}{\columnwidth}
\footnotesize\textit{Note:} All values are percentages. ``---'' indicates that a method is not applicable to the corresponding task.
\end{minipage}
\end{table}

Among the vision-only baselines, YOLOv11 achieves stronger performance than DINO on both applicable tasks. Nevertheless, \method{} improves over YOLOv11 by \(+3.65\%\) in Precision and \(+3.95\%\) in Recall for merging/diverging detection, and by \(+8.24\%\) in P-Acc for drivable-area detection. These results indicate that BEV-image-based detection alone does not fully exploit the upstream geometric priors available for map updating.

The large-model baselines also remain below \method{} on their applicable tasks. This result indicates that vision-language models, even with task-specific fine-tuning or in-context examples, do not satisfy the coordinate-accurate, structured prediction requirements of industrial road-network map updating.

\subsection{Ablation Study}

We evaluate the prior-guided decoder and RGAT, and further isolate the effect of fixing reference points. Table~\ref{tab:ablation} reports the results.

\begin{table}[htbp]
\centering
\caption{Ablation of the prior-guided decoder and RGAT.}
\label{tab:ablation}
\small
\renewcommand{\arraystretch}{1}
\begin{tabular}{
  l
  @{\hspace{2pt}} c
  @{\hspace{8pt}} c @{\hspace{6pt}} c
  @{\hspace{8pt}} c
}
\toprule
\multirow{2}{*}{Variants}
  & \makecell[c]{Intra-\\Intersection}
  & \multicolumn{2}{c}{\makecell[c]{Merging/ \\Diverging}}
  & \makecell[c]{Drivable-\\Area} \\
\cmidrule(lr){2-2} \cmidrule(lr){3-4} \cmidrule(lr){5-5}
  & I-Acc
  & P & R
  & P-Acc \\
\midrule
Ours w/o Prior-Guided Dec. & 92.19 & 91.62 & 88.56 & 95.33 \\
Ours w/o RGAT        & 97.58 & 98.67 & 95.13 & 96.57 \\
Ours w/o Fixed Refs. & 97.75 & 97.51 & 94.25 & 97.71 \\
\midrule
\textbf{Ours}        & \textbf{98.12} & \textbf{98.94} & \textbf{95.64} & \textbf{99.68} \\
\bottomrule
\end{tabular}
\par\smallskip
\begin{minipage}{\columnwidth}
\footnotesize\textit{Note:} ``Ours w/o Prior-Guided Dec.'' replaces the PQM-based query and fixed-reference-point mechanism with learnable queries and iterative reference-point refinement. ``Ours w/o Fixed Refs.'' retains PQM but iteratively refines the geometry-initialized reference points across decoder layers. For merging/diverging detection, ``Ours w/o RGAT'' corresponds to the prior-query model in~\cite{ning2025prior}.
\end{minipage}
\end{table}

\noindent\textbf{Prior-guided decoding.}
Replacing prior-guided decoding with learnable queries and iterative reference-point refinement yields the largest degradation across all tasks. Both variants use the same rendered BEV inputs. The performance gap therefore demonstrates the benefit of encoding geometric priors in decoder queries and reference points, beyond representing geometry in BEV images alone.
\noindent\textbf{RGAT.}
Removing RGAT degrades all tasks, with the most pronounced effect on P-Acc. This suggests that drivable-area decisions depend on relations among map instances and therefore benefit from relational context.
\noindent\textbf{Fixed reference points.}
Iteratively refining geometry-initialized reference points degrades all task metrics. This result supports keeping reference points fixed when reliable element locations are available upstream.

\subsection{Sensitivity Analysis}

% We further analyze how key hyper-parameters and data augmentation affect each task.

All sensitivity analyses in this section are conducted exclusively on the validation set. The held-out city-disjoint test set is not used for model selection and is evaluated only once after all hyperparameters and augmentation settings are fixed. The corresponding test results are reported in Table~\ref{tab:method_comparison}.

% \noindent\textbf{Intra-Intersection Road Network Generation.}
% We vary the maximum number of transit-node candidates \(K_c\) to assess the effect of candidate-space size on topology inference. Table~\ref{tab:sens-kc} shows that \(K_c=20\) achieves higher overall I-Acc than \(K_c=50\). These results indicate that additional candidates enlarge the space of possible road connections without improving overall topology inference. We therefore use \(K_c=20\) in the final model.
\noindent\textbf{Intra-Intersection Road Network Generation.}
We vary the maximum number of transit-node candidates \(K_c\) to assess the effect of candidate-space size on topology inference. Table~\ref{tab:sens-kc} shows that \(K_c=20\) achieves higher overall validation I-Acc than \(K_c=50\). Increasing \(K_c\) enlarges the edge-prediction space but does not improve overall topology inference. We therefore set \(K_c=20\) for the final model.

\begin{table}[htbp]
\centering
\caption{Validation I-Acc under different \(K_c\) values.}
\label{tab:sens-kc}
\small
\setlength{\tabcolsep}{2.8pt}
\renewcommand{\arraystretch}{1}
\begin{tabular}{l r r r r r r r}
\toprule
\# Candidates
& 5--7 & 8--10 & 11--13 & 14--16 & 17--20 & 21--50 & Overall \\
\midrule
\(\boldsymbol{K_c=20}\)
& 98.44 & 97.26 & 96.87 & 96.37 & 97.12 & - & 97.85  \\
\(\boldsymbol{K_c=50}\)
& 98.71 & 97.73 & 95.96 & 94.62 & 96.98 & 94.56 & 97.58 \\
\bottomrule
\end{tabular}
\end{table}

% valid num: 2602， 922，390，204，402，523
%& test result: 98.65 & 97.59 & 96.98 & 96.79 & 97.46 & - &98.12  \\

% \noindent\textbf{Merging/Diverging Detection.}
% Table~\ref{tab:sens-merge} evaluates context size and lane-divider segment length. The results indicate that merging/diverging detection is driven by local geometry: enlarging the context introduces irrelevant lane structures, whereas a \(10\,\mathrm{m}\) divider segment retains sufficient local evidence. We therefore use a context size of 30 and a divider length of \(10\,\mathrm{m}\).
\noindent\textbf{Merging/Diverging Detection.}
Table~\ref{tab:sens-merge} shows that validation performance is
highest with a \(30\,\mathrm{m}\) context and a \(10\,\mathrm{m}\)
divider segment. Larger contexts introduce less relevant lane
structures, confirming that the task primarily relies on local
geometry. We use this configuration in the final model.

\begin{table}[htbp]
\centering
\caption{Validation precision and recall under different BEV context sizes and lane-divider segment lengths.}
\label{tab:sens-merge}
\small
\setlength{\tabcolsep}{8pt}
\renewcommand{\arraystretch}{1.0}
\begin{tabular}{@{}c c c c@{}}
\toprule
Context Size & Divider Len & P & R \\
\midrule
30 & 10\,m & 98.59 & 95.30 \\
30 & 40\,m & 96.28 & 92.82 \\
60 & 10\,m & 70.60 & 81.28 \\
60 & 40\,m & 69.13 & 81.54 \\
\bottomrule
\end{tabular}
\end{table}
%30 & 10\,m & 98.94 & 95.64 \\

% \noindent\textbf{Drivable-Area Detection.}
% Table~\ref{tab:sens-drivable} evaluates training context ranges and rotation augmentation. The results indicate that drivable-area detection benefits from geometric diversity: rotation augmentation improves robustness to orientation variation, while mixed-context training improves robustness to context-scale variation. We therefore use rotation augmentation and mixed-context training.
\noindent\textbf{Drivable-Area Detection.}
Table~\ref{tab:sens-drivable} shows that rotation augmentation improves
validation P-Acc, while mixed-context training improves robustness
across context ranges. We adopt both settings in the final model.

\begin{table}[htbp]
\centering
\caption{Validation P-Acc (\%) under different training context ranges and rotation-augmentation settings.}
\label{tab:sens-drivable}
\small
\setlength{\tabcolsep}{6pt}
\renewcommand{\arraystretch}{1.0}
\begin{tabular}{@{}c c c c@{}}
\toprule
Train Context & Rotation & \makecell[c]{Validation 180--220\,m\\[0pt]} & \makecell[c]{Validation 180--220\,m \&\\60--120\,m} \\
\midrule
180--220\,m & \xmark & 99.50 & 99.28 \\
\addlinespace
180--220\,m & \cmark & 99.63 & 99.43 \\
\addlinespace
\makecell[c]{180--220\,m \&\\60--120\,m} & \cmark & 99.70 & 99.66 \\
\bottomrule
\end{tabular}
\end{table}

\subsection{Real-world Deployment}
\label{sec:deployment}

We deploy the three task-specific pipelines across China in the AutoNavi Map production system.

\noindent\textbf{Production quality monitoring.}
For each task, we assess production quality through manual quality checks of approximately 200 randomly sampled outputs weekly and 500 monthly. We define the task-specific quality score as I-Acc for intra-intersection road network generation, Precision for merging/diverging detection, and P-Acc for drivable-area detection. Over the most recent two months, the corresponding scores are \(98.36\%\), \(98.43\%\), and \(99.80\%\), respectively, all above \(98\%\).

\noindent\textbf{Selective manual verification.}
Before deployment, all production outputs underwent manual verification; after deployment, only cases requiring follow-up are sent for review. The corresponding rates are \(7\%\), \(6\%\), and \(3\%\) for intra-intersection road network generation, merging/diverging detection, and drivable-area detection, respectively. Using the task-specific manual-processing costs and the annual update volume, which covers approximately \(10\%\) of the nationwide road network each year, we estimate the annual manual-processing cost of a fully manual workflow at approximately CNY \(41.0\) million. After deployment, the residual manual-processing cost is approximately CNY \(1.94\) million, corresponding to a \(95\%\) reduction.

\noindent\textbf{Update frequency.}
Table~\ref{tab:deployment-efficiency} reports their nationwide inference efficiency under ten-GPU parallelism. The combination of nationwide inference efficiency and selective manual verification increases the map-update frequency from weekly to hourly.

\begin{table}[htbp]
\centering
\small
\setlength{\tabcolsep}{6pt}
\renewcommand{\arraystretch}{1.0}
\caption{Deployment statistics across China for the three update tasks.}
\label{tab:deployment-efficiency}
\begin{tabular}{@{}c c c c@{}}
\toprule
Task & \makecell[c]{Nationwide\\Coverage} & \makecell[c]{Time\\(ten GPUs)} & \makecell[c]{FPS\\(per GPU)} \\
\midrule
\makecell[c]{Intra-\\Intersection}
& 2.31M intersections & 6.83 h & 9.4 \\
\addlinespace
\makecell[c]{Merging/\\Diverging}
& \makecell[c]{1.26M candidate instances\\(0.72M locations detected)}
& 0.63 h & 55.6 \\
\addlinespace
\multirow{2}{*}{\makecell[c]{Drivable-\\Area}}
& 2.31M intersections & 4.06 h & 15.8 \\
& 9.24M road segments & 8.12 h & 31.6 \\
\bottomrule
\end{tabular}
\end{table}
% 4.06，8.12

\section{Conclusion}

We presented \method{}, a prior-guided unified decoding framework for industrial road-network map updating. By preserving upstream geometric priors during decoding and modeling dependencies among related instances, it provides a shared architecture instantiated separately for three update tasks. \method{} outperforms all applicable baselines in offline experiments. Following deployment in the AutoNavi Map production system, it achieves quality scores above \(98\%\), reduces manual verification costs by \(95\%\), and increases update frequency from weekly to hourly.

The results indicate that, when reliable geometry is available upstream, preserving it as a decoding prior is more effective than re-estimating locations during decoding. The framework remains dependent on upstream geometry and has been evaluated under AutoNavi Map production specifications in China. Future work will extend the framework to additional map-update tasks, investigate uncertainty-aware decoding for incomplete or erroneous upstream geometries, and evaluate transfer across map sources and production specifications.

\bibliographystyle{ACM-Reference-Format}
\bibliography{refs}

\appendix

\section{Dataset Statistics}
\label{sec:appendix-datasets}

% Table~\ref{tab:dataset_stats} summarizes the statistics of the three
% task-specific datasets. The three tasks operate on road-level, coarse
% lane-level, and high-quality lane-level map data, respectively, and each dataset
% uses a task-specific sampling unit and urban-to-suburban ratio. All training,
% validation, and test splits are drawn from spatially disjoint map tiles to
% prevent geographic leakage.

Table~\ref{tab:dataset_stats} summarizes the three task-specific datasets. The tasks operate on road-level, coarse lane-level, and high-quality lane-level map data, respectively, with task-specific sampling units and urban-to-suburban ratios. The source cities for training and validation are Beijing, Guangzhou, Shenzhen, Nanjing, Dongguan, Huizhou, Suzhou, and Wuhan. Shanghai, Hangzhou, Chongqing, and Langfang are held out exclusively for testing.

\begin{table}[htbp]
\centering
\small
\renewcommand{\arraystretch}{1.25}
\setlength{\tabcolsep}{5pt}
\caption{Statistics of the task-specific datasets.}
\label{tab:dataset_stats}
\begin{tabular}{@{}c c c c c@{}}
\toprule
Task & Level & Unit & \# Samples & Urban\,:\,Sub. \\
\midrule
\makecell[c]{Intra-\\Intersection} & Road   & Intersection & 125{,}929 & 7\,:\,3 \\
\addlinespace[4pt]
\makecell[c]{Merging/\\Diverging}  & Coarse & Divider pair & 25{,}720 & 8\,:\,2 \\
\addlinespace[4pt]
\makecell[c]{Drivable-\\Area}      & Fine   & Intersection & 42{,}716 & 6\,:\,4 \\
\bottomrule
\end{tabular}
\par\smallskip
% \begin{minipage}{\columnwidth}
% \footnotesize\textit{Note:} Road, Coarse, and Fine denote road-level, coarse lane-level, and high-quality lane-level map data, respectively; Sub.\ denotes suburban samples. ``Divider pair'' denotes a retrieved pair of lane dividers. The training, validation, and test splits use spatially disjoint map tiles in an \(8:1:1\) ratio. For drivable-area detection, intersections are sampled because complex cases require model-based processing, while simple road-segment cases are handled by production rules.
% \end{minipage}
\begin{minipage}{\columnwidth}
\footnotesize\textit{Note:} Road, Coarse, and Fine denote road-level, coarse lane-level, and high-quality lane-level map data, respectively; Sub.\ denotes suburban samples. ``Divider pair'' denotes a retrieved pair of lane dividers. For drivable-area detection, intersections are sampled because complex cases require model-based processing, while simple road-segment cases are handled by production rules.
\end{minipage}
\end{table}

\section{Scenario-Level Analysis}
\begin{figure}[!htbp]
\centering
\includegraphics[width=0.8\linewidth]{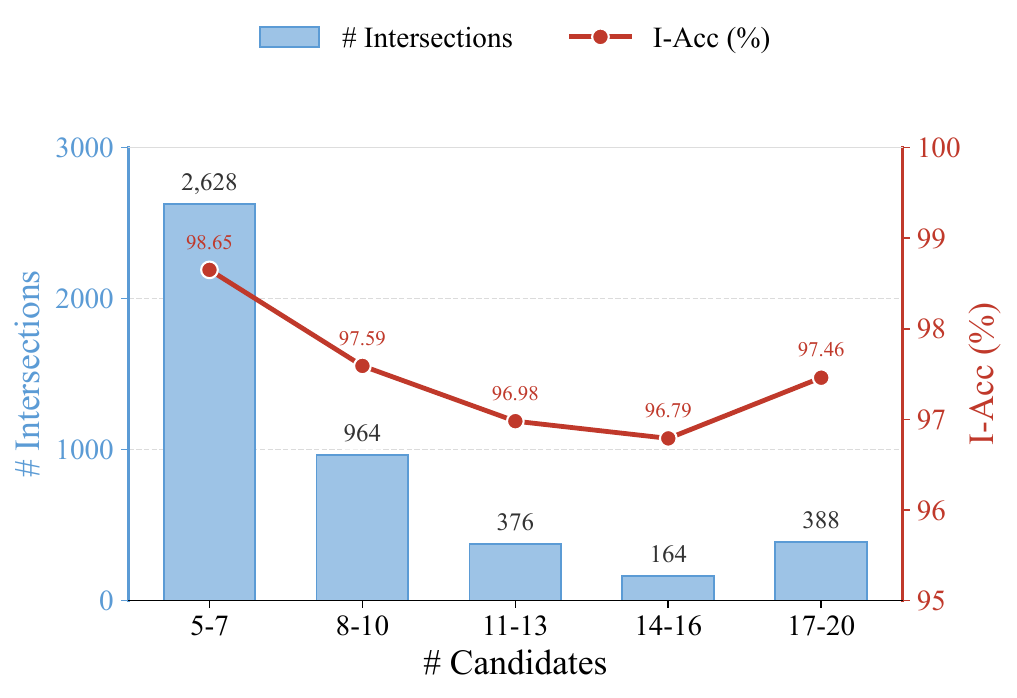}
\caption{I-Acc by node-set size.}
\label{fig:main_res_a}
\end{figure}

Figure~\ref{fig:main_res_a} shows that larger candidate sets expand the edge-prediction space and are associated with modestly lower I-Acc. Despite this increase in complexity, performance remains consistently high; the non-monotonic trend indicates that candidate count alone does not determine intersection complexity.

\begin{figure}[!htbp]
\centering
\includegraphics[width=0.8\linewidth]{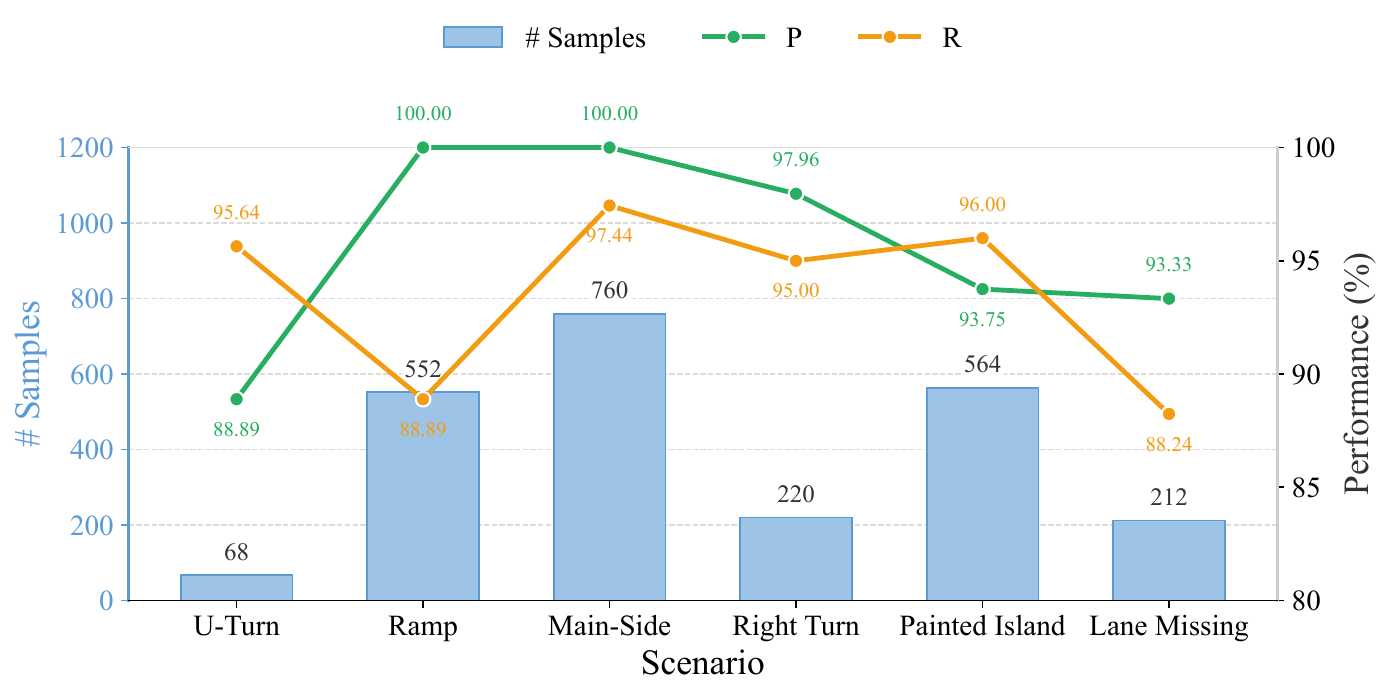}
\caption{Merging/diverging performance by scenario.}
\label{fig:main_res_b}
\end{figure}

Figure~\ref{fig:main_res_b} reveals distinct error profiles across scenarios: lane-missing cases have the lowest Recall, consistent with incomplete lane-divider evidence, whereas U-turn cases have the lowest Precision, reflecting greater geometric ambiguity.

\section{Task-Specific Relation Types}
\label{app:relation_types}

For each task, we define seven valid relation types (\(R=7\)). 
Accordingly,
\[
M_{ik}^{\mathrm{rel}}\in\{0,1,\ldots,R\},
\]
where \(M_{ik}^{\mathrm{rel}}=0\) indicates that no valid relation exists,
and the corresponding instance pair is masked during attention.

As illustrative examples, \(M_{ik}^{\mathrm{rel}}=1\) denotes that
instances \(i\) and \(k\) are directly connected in the upstream map,
whereas \(M_{ik}^{\mathrm{rel}}=2\) denotes that they are not directly
connected but have a geometric distance of at most \(5\,\mathrm{m}\).
The remaining labels encode other task-specific relations.

\section{Qualitative Examples}
\label{sec:appendix-examples}

\begin{figure}[!htbp]
    \centering
    \includegraphics[width=0.8\linewidth]{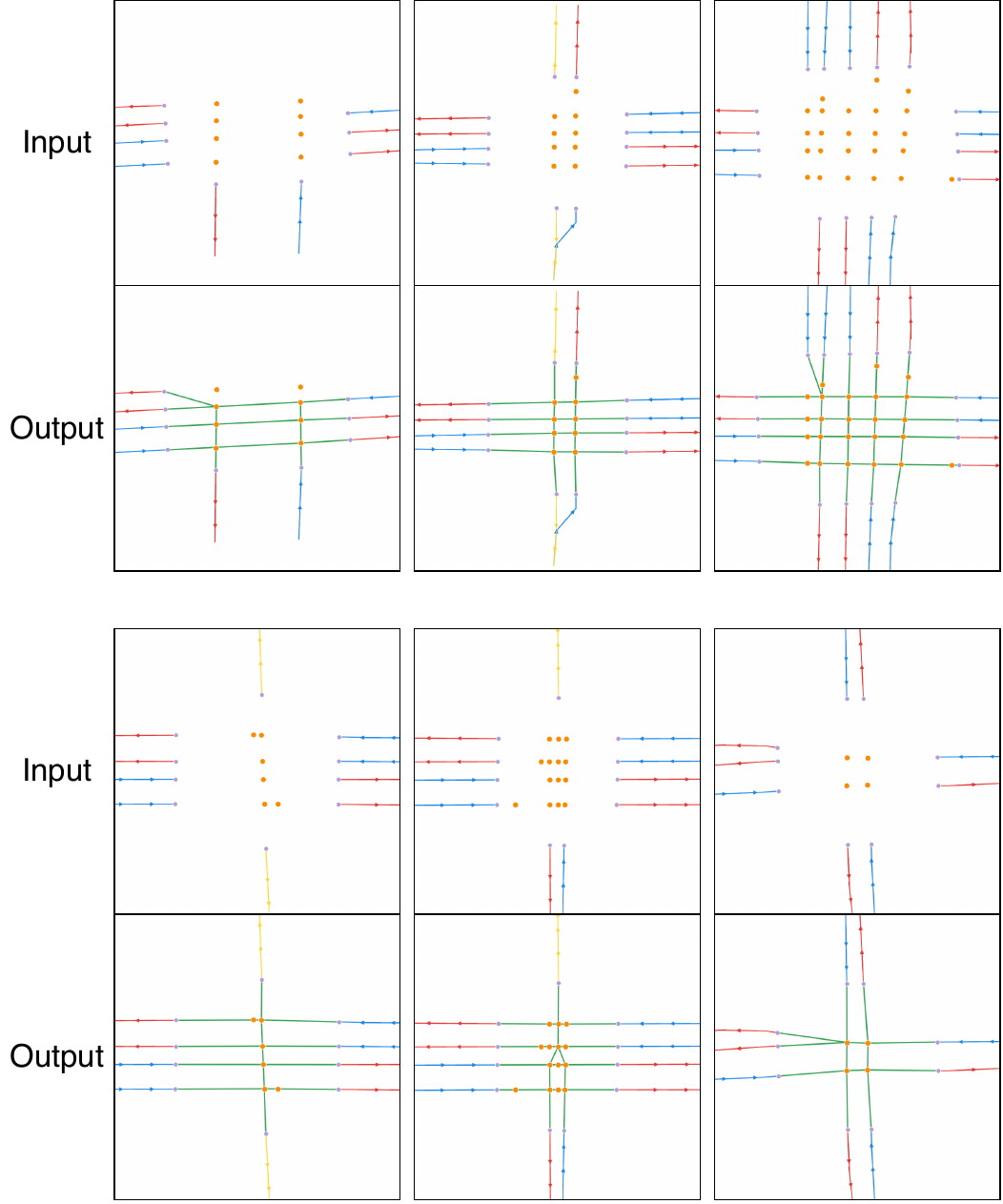}
    \caption{Qualitative results for intra-intersection road
    network generation.}
    \label{fig:rendering_a}
\end{figure}

\begin{figure}[!htbp]
    \centering
    \includegraphics[width=0.8\linewidth]{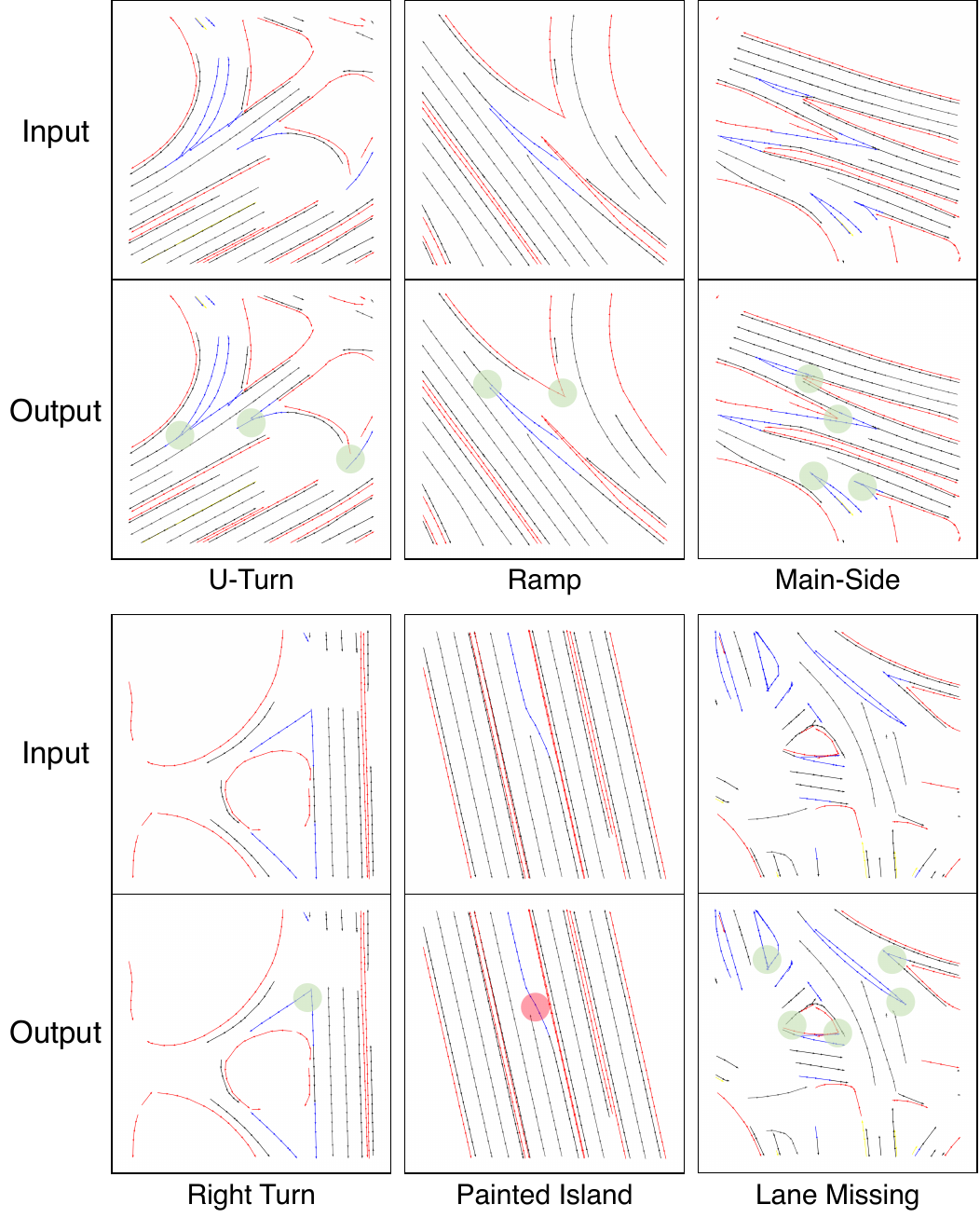}
    \caption{Qualitative results for merging/diverging
    detection.}
    \label{fig:rendering_b}
\end{figure}

\begin{figure}[!htbp]
    \centering
    \includegraphics[width=0.8\linewidth]{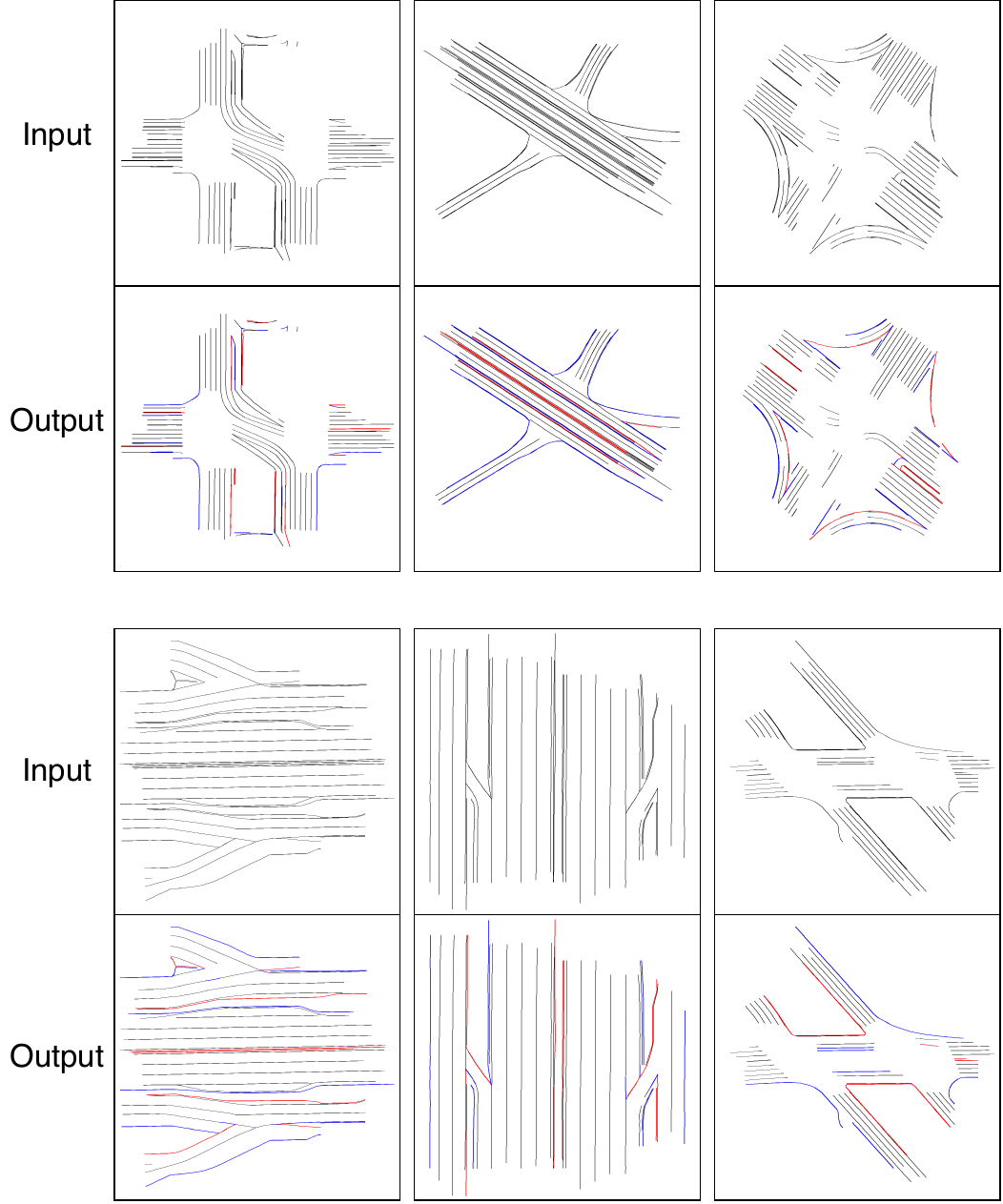}
    \caption{Qualitative results for drivable-area detection.}
    \label{fig:rendering_c}
\end{figure}

To complement the quantitative evaluation, we visualize
input--output pairs for each of the three road-network map
updating tasks in Figures~\ref{fig:rendering_a}--\ref{fig:rendering_c}.
Each figure contains six representative cases drawn from the
production test set, arranged in four rows: rows~1 and~3
display the upstream map geometry serving as model input, and
rows~2 and~4 display the corresponding prediction overlaid on
the same region. Color conventions follow
Figure~\ref{fig:illustration}.

\section{Failure Case Analysis}
\label{sec:appendix-failure}

\begin{figure}[!htbp]
    \centering
    \includegraphics[width=0.8\linewidth]{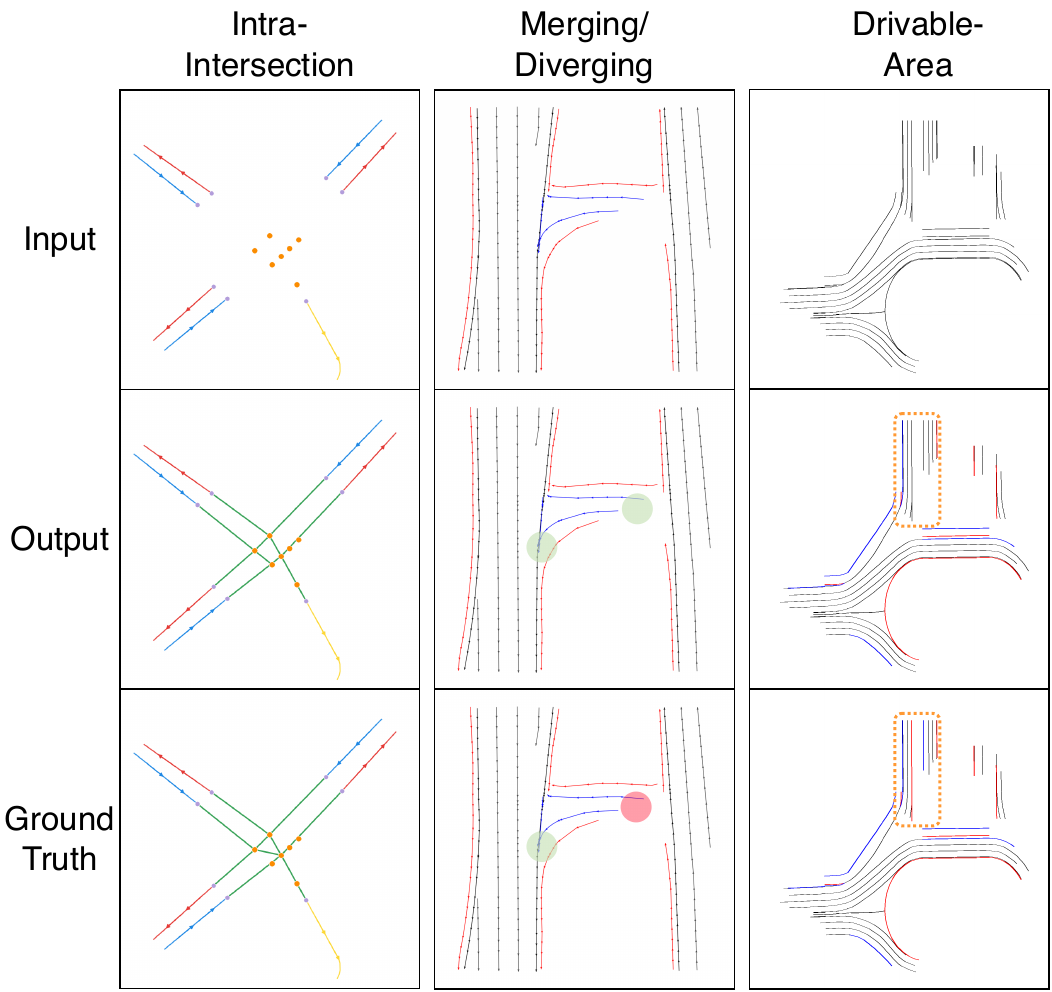}
    \caption{Representative failure cases of \method{} on the three
    tasks.}
    \label{fig:bad_case}
\end{figure}

Figure~\ref{fig:bad_case} presents one representative failure case for
each of the three tasks. The three columns correspond to
intra-intersection road network generation, merging/diverging
detection, and drivable-area detection; the three rows show the
upstream map geometry used as input, the prediction produced by
\method{}, and the ground truth, respectively. Color conventions
follow Figure~\ref{fig:illustration}.

% \noindent\textbf{Intra-Intersection Road Network Generation.}
% In this case, \method{} correctly recovers all ten ground-truth
% connections but additionally predicts one connection in the
% skew-crossing conflict region that does not exist in the ground truth
% (one false positive), a mild case of over-connection. Such errors typically arise in intersections with irregular branch geometry, where several candidate connections are geometrically plausible.

\clearpage
\end{document}